\documentclass[letterpaper,10pt,conference]{ieeeconf}
\IEEEoverridecommandlockouts
\usepackage{amsmath,amssymb}
\usepackage{booktabs}
\usepackage{graphicx}
\usepackage{url}
\usepackage{balance}
\usepackage[T1]{fontenc}
\usepackage[utf8]{inputenc}

\usepackage{cuted}
\usepackage{capt-of}

\title{\LARGE \bf
SocialVLA: A Social Perception Gateway for Human-Reaction-Based\\
Failure Detection and Recovery in VLA Manipulation
}

\author{
Sofya Konstantinova$^{1,2}$,
Miguel Altamirano Cabrera$^{1,2}$, \\
Artem Lykov$^{2,1}$,
and Dzmitry Tsetserukou$^{1}$
\thanks{$^{1}$Intelligent Space Robotics Lab, Skolkovo Institute of Science and Technology, Moscow, Russian Federation.}
\thanks{$^{2}$R\&D Center, MWS, Moscow, Russian Federation.}
\thanks{E-mails: \{sofya.Konstantinova, m.altamirano, artem.lykov, d.tsetserukou\}@skoltech.ru.}
}

\begin{document}
\bstctlcite{IEEEexample:BSTcontrol}

\maketitle
\begin{strip}
\centering
\vspace{-8mm}

\includegraphics[width=0.96\textwidth]{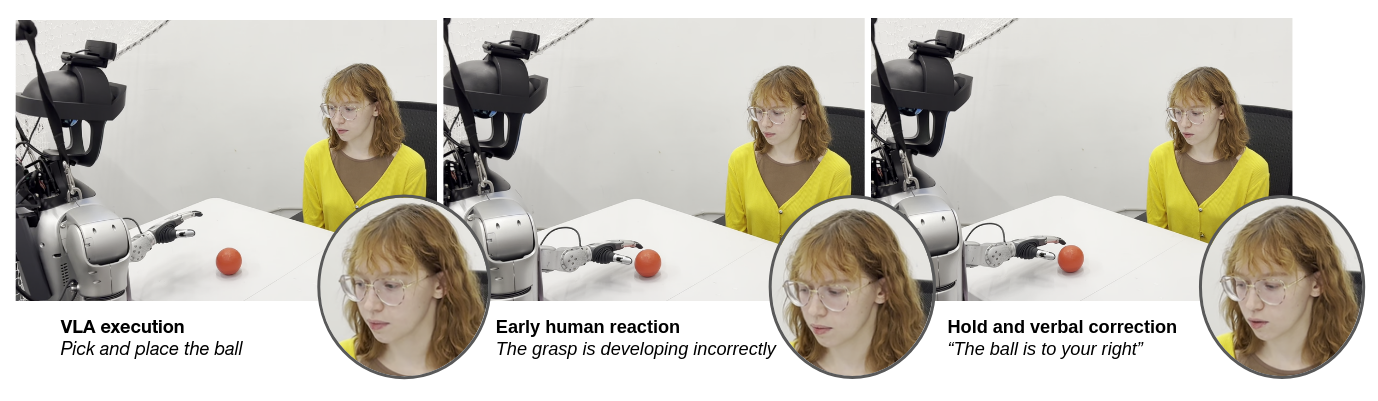}
\captionof{figure}{We introduce SocialVLA, a local social-perception gateway that uses spontaneous human reactions---including speech, vocal affect, and facial behavior---to detect emerging VLA manipulation errors in real time, request a VLA hold, and record the participant's verbal correction.}
\label{fig:gasp_overview}
\end{strip}

\begin{abstract}

Vision-language-action (VLA) policies enable diverse robotic manipulation but remain susceptible to execution failures that may not be recognized by the policy itself. An observing human provides complementary information: unexpected robot behavior can elicit rapid vocal, facial, or explicit verbal reactions before an undesirable manipulation outcome is completed. We introduce SocialVLA, a local and policy-agnostic social perception gateway that converts spontaneous human reactions into runtime intervention signals for VLA manipulation. SocialVLA combines causal paralinguistic audio detection, visual reaction recognition, explicit stop phrases, and robot-relevance estimation. An asynchronous first-event fusion mechanism allows the earliest sufficiently confident signal to request a VLA hold without waiting for complete multimodal agreement. After intervention, a separate speech channel records the participant's verbal correction and associates it with the interrupted VLA action, enabling participant-directed continuation, restart, or instruction revision. We evaluate SocialVLA during physical Unitree G1 manipulation using recordings from 15 participants, comprising 238 annotated intervention-worthy reaction episodes and 1.038~h of annotated non-intervention behavior. Frozen offline replay of the multimodal cascade achieves 54.6\% event recall and 69.5\% precision, while unfiltered first-event audio--video fusion reaches 64.3\% recall. The relevance module reduces false-stop episodes from 100 to 57 and increases precision from 60.5\% to 69.8\%. In a prospective deployment on a 16th unseen participant, the frozen system achieves 59.5\% recall and 91.7\% precision. Median detector-to-fusion delivery is 47.9~ms, median VLA-gate-to-physical-hold latency is 336~ms, and matched reaction-onset-to-physical-hold latency is 1.021~s. These results demonstrate a complete local pathway from spontaneous social reaction to physical VLA interruption and a participant-directed recovery interface.

\end{abstract}

\noindent\textbf{Index Terms---}Failure Detection and Recovery, Multi-Modal Perception for HRI, Human Factors and Human-in-the-Loop.

\section{INTRODUCTION}\label{introduction}

Vision-language-action (VLA) models map visual observations and language instructions to robot actions, enabling policies to generalize across objects, tasks, and semantic commands \cite{ref1,ref2}. Their generality, however, does not eliminate execution failures. During pick-and-place manipulation, a policy can select the wrong object, generate an unsuitable approach, fail to establish a grasp, drop an object, or place it at an incorrect location. These failures may occur even when individual actions appear locally plausible. 

Recent methods detect failures from robot observations, trajectories, uncertainty estimates, or internal VLA representations. SAFE learns a failure score from VLA features and successful and failed rollouts \cite{ref3}. FAIL-Detect frames runtime monitoring as sequential out-of-distribution detection \cite{ref4}. AHA uses a vision-language model to identify and explain manipulation failures \cite{ref5}. These approaches provide direct task evidence, but they may require access to policy internals, compatible observations, representative training data, or an explicit model of task progress.

An observing person provides a different source of information. Humans can discriminate complex visual scenes within a fraction of a second \cite{ref6}, while nonverbal vocalizations can communicate recognizable affect within several hundred milliseconds \cite{ref7}. When a robot acts unexpectedly, a person may use an explicit stop phrase, alter vocal prosody, display a negative facial reaction, shake their head, or redirect attention. These responses reflect the person's appraisal of the interaction \cite{ref8} and may reveal discrepancies that were not enumerated during policy training. Moreover, an observer need not wait until a failure is physically complete. From the evolving scene, motion direction, and task context, a person may recognize that a grasp or placement is developing incorrectly and react before the object is dropped, displaced, or released at the wrong location. Recent HRI benchmarks explicitly distinguish such anticipatory responses from reactions observed after failure completion \cite{ref18}.

Prior human-robot interaction (HRI) studies have established that human reactions correlate with robot errors \cite{ref9,ref10,ref11,ref12,ref13,ref14}. However, retrospective reaction classification is not equivalent to an operational gate for a continuously executing VLA. In manipulation, detection time is part of the task: a correct decision made after an object has already been dropped or incorrectly placed cannot prevent the consequence. A runtime system must therefore process the earliest reliable signal with low delay, handle missing or uncertain modalities, and avoid waiting for unnecessary multimodal agreement. It should run locally without dependence on an Internet service, remain independent of the internal VLA architecture, and connect as an external layer to any policy that exposes a pause, hold, or stop interface. Repeated evidence must reinforce an intervention rather than reverse an already issued stop. The key challenge is therefore to convert spontaneous social behavior into a sufficiently fast and reliable runtime intervention signal for a continuously executing VLA, while avoiding unnecessary stops caused by unrelated human behavior.


We address this challenge with SocialVLA, a modular social perception gateway between a human observer and an executing VLA. Rather than attempting to infer physical failure solely from the robot state, SocialVLA detects human-perceived discrepancies from continuous audio-visual behavior and converts them into structured runtime intervention events. The framework is local and does not require access to VLA activations, action probabilities, or internal uncertainty estimates. In this work, failure detection refers to detecting human-perceived intervention need during robot execution rather than exhaustively identifying all physical robot failures.

The main contributions are:

\begin{enumerate}
\def\labelenumi{\arabic{enumi})}
\item
  SocialVLA, a local and policy-agnostic social perception gateway that converts
  spontaneous human audio-visual reactions and explicit stop commands into
runtime interventions during VLA-controlled humanoid manipulation;

\item
a causal multimodal intervention architecture that combines fast and
extended-context paralinguistic detection, visual reaction recognition,
explicit stop phrases, separately estimated robot relevance, and
asynchronous first-event fusion, allowing the earliest sufficiently
confident signal to request a hold without waiting for complete multimodal
agreement; and

\item
a real-robot evaluation on the Unitree G1 using a 15-participant annotated
corpus and a prospective frozen-system deployment on a 16th, previously
unseen participant, measuring reaction detection, false interventions,
timely recall, and end-to-end physical stopping latency.
\end{enumerate}


Together, these contributions establish spontaneous social behavior as an operational input to VLA execution rather than only an offline observation signal. SocialVLA complements robot-internal failure monitors by contributing the observer's task-level assessment while remaining independent of the internal architecture of the underlying VLA.

\section{RELATED WORK}\label{related-work}

\subsection{Failure Detection for Robot Policies}\label{failure-detection-for-robot-policies}

VLA policies extend vision-language representations to robotic action generation \cite{ref1,ref2}. SAFE estimates task failure using internal VLA features and studies the trade-off between detection accuracy and timeliness \cite{ref3}. FAIL-Detect extracts uncertainty-related signals from policy inputs and outputs and applies conformal sequential detection \cite{ref4}. AHA instead observes robot trajectories and produces semantic failure judgments and explanations \cite{ref5}. 

SocialVLA is complementary to these methods. It does not require VLA activations, action probabilities, uncertainty estimates, or a task-specific success predicate. The same external reaction channel can therefore be attached to different policies. The cost of this independence is that SocialVLA cannot verify whether the observer's interpretation is physically correct. A practical robot should combine human-centered and robot-internal evidence rather than treating either source as universally reliable.

\subsection{Human Reactions to Robot Errors}\label{human-reactions-to-robot-errors}


Stiber et al.~showed that multimodal human responses can support timely robot-error detection \cite{ref9}. The BAD dataset subsequently captured bystander reactions to recorded robot failures \cite{ref10}, while REFLEX studied multimodal reactions to robot failures and subsequent explanations \cite{ref12}. ERR@HRI introduced standardized multimodal benchmarks and time-sensitive evaluation of error-related human behavior \cite{ref13,ref18}. Recent observations further show that human reactions remain informative in less controlled environments, although unrelated stimuli, groups, and repeated failures introduce substantial noise \cite{ref14}.

Most closely related to SocialVLA, Stiber et al.~implemented a proactive system that combines facial behavior, speech, explicit feedback, and robot-error context during live interaction \cite{ref11}. SocialVLA differs in its intended role as an external gateway around a continuously executing VLA: it does not require access to VLA internals, explicitly separates reaction recognition from robot-relevance estimation, and prioritizes causal first-event intervention so that a sufficiently strong unimodal signal need not wait for multimodal agreement. The objective is therefore not only to recognize a human response retrospectively, but to convert that response into a actionable hold request during physical VLA execution.


\subsection{Human Reactions as Temporal Supervision}\label{human-reactions-as-temporal-supervision}

Beyond triggering intervention, a detected human response indicates that the recent execution has become inconsistent with the observer's expected task outcome. SocialVLA records this evidence without requiring the observer to operate a dedicated control device. The synchronized stop record identifies the active VLA action or chunk and associates it with the subsequent verbal correction. Such synchronized reaction and correction records may provide temporal supervision for future failure-aware or preference-based policy learning.

\section{SYSTEM ARCHITECTURE}\label{system-architecture}

\subsection{Event Representation and
Scope}\label{event-representation-and-scope}

Let $x_a(t)$ and $x_v(t)$ denote participant-facing audio and video streams. A detector produces:

\begin{equation} e_i=(m_i,c_i,s_i,[t_i^0,t_i^1],t_i^e), \label{eq:event} \end{equation}
where $m_i$ is the modality, $c_i$ the event class, $s_i$ the score, $[t_i^0,t_i^1]$ the source interval supporting the decision, and $t_i^e$ the publication time. The system distinguishes fast reaction events, extended-context reaction events, and direct lexical events. Attention-target estimates are diagnostic; only the binary robot-relevance probability enters the candidate fusion rule.

\begin{figure}[t]
\vspace*{6pt}
\centering
\includegraphics[width=0.95\columnwidth]{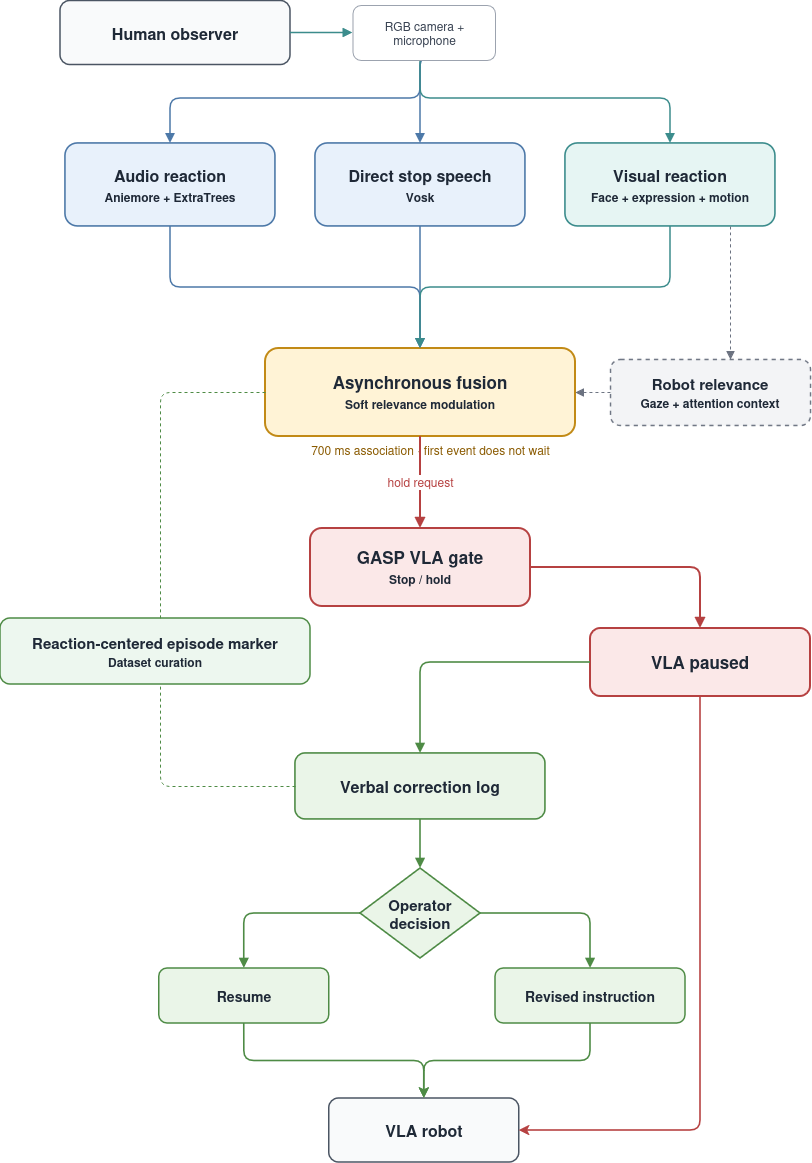}
\caption{SocialVLA architecture. Independent audio, lexical, and visual reaction
channels produce asynchronous evidence. The selected soft fusion rule combines
reaction confidence with bounded robot-relevance context. The first authorized
event requests a VLA hold, while subsequent evidence is attached to the open
reaction episode. A post-stop speech log records the participant's correction
without affecting the robot hold state.}
\label{fig:architecture}
\end{figure}

\subsection{Audio Reaction Paths}\label{audio-reaction-paths}

Audio is recorded at 16 kHz in mono. Each base observation is a causal 320-ms window. The Aniemore emotion model \cite{ref19}, based on wav2vec~2.0 \cite{ref15}, produces seven emotion logits that form the acoustic representation used by the downstream reaction heads.

For a causal sequence of seven-dimensional observations $H_t$ within a history $h$, the temporal representation is:

\begin{equation} \phi_h(t)=[z_t,\operatorname{mean}(H_t),\operatorname{std}(H_t),z_t-z_{\mathrm{first}}], \label{eq:audiofeatures} \end{equation}
where $z_t$ is the latest emotion-logit vector. Equation~\eqref{eq:audiofeatures} produces 28 features. A gap longer than 0.21 s clears the history so that unrelated audio segments are not joined.

The fast head uses $h\leq0.4$~s and an ExtraTrees classifier with 160 trees and minimum leaf size 2. The extended-context head uses the same feature definition but $h\leq2.0$ s and is trained independently with the same ExtraTrees hyperparameters. The fast path is intended to detect a reaction from the earliest available acoustic dynamics. The extended-context path captures prosodic evidence that develops beyond the fast window. Both heads belong to the online intervention stage; the extended-context estimate can enrich an open reaction episode but is not a prerequisite for issuing an earlier authorized hold request.

Audio supervision is strictly modality-local. Its label specifies whether the acoustic behavior contains an intervention-worthy reaction. The fields \texttt{robot\_related} and \texttt{attention\_target} are not used as audio inputs, audio targets, sample-selection conditions, or sample weights. This prevents a model trained on speech from acquiring target information available only from synchronized video annotation.

\subsection{Direct Lexical Path}\label{direct-lexical-path}

Vosk remains an independent fast channel for authorized Russian stop commands and negative phrases. It does not replace the learned audio heads. An accepted Vosk event contributes a unit-strength lexical candidate. Direct stop phrases can therefore trigger the cascade even when vocal affect is weak, while the Aniemore heads can detect nonlexical or prosodic reactions before complete transcription.

\subsection{Video Reaction Head}\label{video-reaction-head}

YuNet \cite{ref16} detects the participant's face, and five landmarks are used to align it. A MobileFaceNet facial-expression model produces seven facial-expression scores. A 25-dimensional per-frame vector contains these scores, normalized five-landmark geometry, face-detector confidence, normalized face position and size, and normalized yaw, pitch, and roll. The vector is concatenated with its causal 400-ms difference, producing 50 video-reaction features.

An ExtraTrees classifier with 160 trees and minimum leaf size 20 maps these features to $p_v$, the probability of an intervention-worthy visible reaction. Horizontally mirrored video features augment only the training partitions and the final candidate fit; mirrored samples are never added to a held-out fold.

\subsection{Robot-Relevance and Attention
Heads}\label{robot-relevance-and-attention-heads}

Robot relevance is learned by a separate video-only binary logistic head. Its inputs combine gaze location, face availability, attention confidence, estimated attention states, selected object-detection scores, and facial features. The head produces: 
\begin{equation} r=P(\mathrm{robot\_related}\mid x_v),\qquad r\in[0,1]. \label{eq:relevance} \end{equation}

The relevance target comes exclusively from video annotation.

A four-class attention head estimates robot camera, robot workspace, external object, or away. Low-confidence predictions are rejected as unknown. This categorical output remains diagnostic and is not a hard stop gate. In particular, an estimated away direction does not act as an independent
veto; the intervention decision remains determined by the fused confidence
and operating threshold. 

\subsection{Soft Multimodal Fusion}\label{soft-multimodal-fusion}

Let $p_a^f$ and $p_a^c$ denote the fast and extended-context audio probabilities, $p_v$ the video-reaction probability, and $v\in\{0,1\}$ an authorized Vosk event. For stage $q\in\{f,c\}$, the raw reaction evidence is: 
\begin{equation} u_q=\max(p_a^q,p_v,v). \label{eq:rawfusion} \end{equation}

The relevance factor depends on whether the evidence includes a direct lexical stop event:

\begin{equation}
a(r,v)=\begin{cases}
0.85+0.15r, & v=1,\\
0.70+0.30r, & v=0,
\end{cases}
\label{eq:factor}
\end{equation}
and the stage confidence is:
\begin{equation} s_q=u_q a(r,v). \label{eq:score} \end{equation}

For an explicit stop phrase, the 0.85 floor keeps a direct human command above the selected threshold even when visual relevance is uncertain. For prosodic and facial evidence, relevance can reduce confidence by at most 30\%. It therefore remains a contextual cue rather than a veto. If no recent relevance observation is available, the runtime uses $r=0.5$.

The selected operating point is $\theta=0.772$. The fast and
extended-context paths are evaluated independently, and the first authorized
threshold crossing opens the intervention episode. The selected fusion rule,
threshold, and temporal-association parameters are fixed after cross-validated
development-set selection and remain unchanged during the prospective
evaluation.

\subsection{Temporal Association and First-Event Intervention}\label{temporal-association-and-first-event-intervention}

The latest audio, video, Vosk, and relevance observations are associated when their source times fall within $W=700$ ms. This interval accommodates natural cross-modal asynchrony and unequal inference times. It is an association window, not an additional waiting period.

At deployment, the first authorized candidate may open a stop episode immediately. Later extended-context and cross-modal evidence are attached to the same episode. The intervention time is therefore:
\begin{equation} t_{\mathrm{request}}=\min_i\{t_i^e:c_i\in\mathcal{A}\}. \label{eq:request} \end{equation}

Here $\mathcal{A}$ contains the authorized fast, extended-context, and direct lexical stop-event classes.

This policy differs from audio-AND-video fusion, which must wait until both modalities are available and can miss unimodal reactions.

\subsection{Stop-and-Correct Interface}\label{stop-and-correct-interface}

When SocialVLA detects an intervention-worthy reaction, the \texttt{gasp-vla-gate} sends one pause command to the running VLA and places it in a hold state. Further reaction events belonging to the same intervention episode do not produce additional control commands. The robot remains paused until the participant explicitly requests continuation. The participant may resume execution from the held state, adjust the workspace and restart the action, or provide a revised instruction for a subsequent VLA action.

The language interface has two distinct roles. During execution, the low-latency Vosk path recognizes a restricted vocabulary of explicit stop commands and negative phrases. After the hold request, a separate Vosk logging session records unrestricted verbal correction together with its timestamp. This post-stop transcript is non-safety-critical: it cannot resume the robot, clear the hold state, or issue motion commands. The participant retains exclusive authority to continue execution through the explicit resume command. Separating immediate lexical interruption from post-stop transcription avoids delaying the first hold request.

\subsection{Reaction-Centered Intervention Records}\label{reaction-centered-markers}

For every intervention $k$, SocialVLA stores an episode record:
\begin{equation}
M_k=(id_k,t_r,t_s,t_c,a_k,\mathcal{M}_k,\mathcal{E}_k,y_k),
\label{eq:marker}
\end{equation} 
where $t_r$, $t_s$, and $t_c$ denote the synchronized reaction, stop-request, and correction times; $a_k$ identifies the active VLA action or chunk; $\mathcal{M}_k$ and $\mathcal{E}_k$ store the contributing modalities and detector evidence; and $y_k$ stores the correction transcript. The record therefore associates the observer's reaction and verbal correction with the VLA action active at the time of intervention. These records support analysis of human-identified problematic execution segments and may provide supervision for subsequent failure-aware or preference-based policy-data curation; policy retraining itself is outside the scope of this work.


\section{EXPERIMENTAL METHODOLOGY}\label{experimental-methodology}

\subsection{Platform and Task}\label{platform-and-task}


Experiments use a physical Unitree G1 humanoid controlled by GR00T-WholeBodyControl and a GR00T policy trained for a pick-and-place task. The robot must grasp a ball and place it in a specified target region. Participants observed both successful executions and naturally occurring VLA failures. Failure occurrence was not announced, and participants were not instructed to produce specific reactions.

The prediction target in this study is the observer's intervention-worthy response rather than a predefined taxonomy of physical failure modes. Consequently, the evaluation asks whether SocialVLA can recognize when a human observer perceives the ongoing execution as requiring intervention and can translate that perception into a timely VLA hold.

\subsection{Recordings and Annotation}\label{recordings-and-annotation}

We collected capture-clock-synchronized audio--video timeline data across 15 finalized sessions from 15 distinct participants. Each participant stood approximately 1 m from the physical robot, could see both the robot and its workspace, could hear its operation, and knew that the task was to grasp and place the ball. Each session contains participant-facing RGB video and microphone audio synchronized with robot and model logs. 


Intervals were labelled by modality, intervention status (intervention-worthy or non-intervention), robot-related status, attention target, reaction subtype, and an event identifier linking modalities. Reaction onset is the first instant at which an intervention-worthy cue is recognizable from the isolated short clip; offset is the instant at which that cue disappears. Robot-error markers are recorded on the synchronized robot timeline, while detection latency is measured from human reaction onset. Only finalized, synchronization-verified annotations are used. Unannotated time is excluded and is never automatically labelled as non-intervention behavior.

The finalized multimodal annotations contain 238 robot-related intervention-worthy reaction episodes and 62.3~min of explicitly annotated non-intervention behavior. Audio and video manifestations that overlap or are separated by no more than 700~ms are automatically merged into one reaction episode.


The positive class comprises observable reactions indicating that the participant perceives the ongoing robot execution as requiring intervention. These include explicit stop commands and spontaneous vocal, facial, or gestural responses. The target therefore represents human-perceived intervention need rather than an exhaustive ground truth over all physical robot VLA failure modes.

\subsection{Evaluation Protocol}
\label{evaluation-protocol}

We use two complementary evaluation settings. First, the frozen multimodal
cascade and its ablations are replayed over the annotated corpus collected
from 15 participants without retraining or per-ablation threshold selection.
This full-corpus analysis measures channel complementarity and operating
trade-offs but is not presented as an unseen-participant generalization
estimate. Second, the complete frozen system is evaluated prospectively on
a 16th, previously unseen participant during live VLA execution. This
prospective participant is excluded from model fitting, hyperparameter
selection, threshold selection, and calibration.

The primary evaluation unit is one annotated reaction episode and one predicted stop episode. Intervention-worthy audio and video intervals form one reaction episode when they overlap or are separated by at most 700 ms. A predicted episode begins at the first threshold crossing; consecutive crossings separated by no more than 400 ms belong to the same prediction. One-to-one matching prevents a prolonged reaction from contributing multiple true positives. A prediction matches a reaction only when its first threshold crossing occurs between the annotated onset and 1 s after onset. The primary timely criterion uses the stricter interval from onset to 500 ms. Predictions preceding the annotated human-reaction onset are not counted as reaction detections.

The development operating point maximizes event recall subject to an exploratory non-intervention false-positive-rate target of 0.10. Hyperparameters and thresholds are selected using development data and then frozen before full-corpus replay and prospective deployment. The 700-ms association interval is fixed as a cross-modal event-linking tolerance and is not added to detection latency.

\subsection{Metrics}\label{metrics}

For event predictions, we report:

\begin{equation} \mathrm{Precision}=\frac{TP}{TP+FP}. \label{eq:precision} \end{equation}

\begin{equation} \mathrm{Recall}=\frac{TP}{TP+FN}. \label{eq:recall} \end{equation}

\begin{equation}
\mathrm{NonInt\mbox{-}FPR}
=
\frac{N_{\mathrm{NonInt,triggered}}}
     {N_{\mathrm{NonInt}}},
\label{eq:fpr}
\end{equation}
here, $N_{\mathrm{NonInt}}$ is the number of manually delimited non-intervention or robot-unrelated episodes, and $N_{\mathrm{NonInt,triggered}}$ is the number of those episodes containing at least one predicted stop. Because multiple false stop episodes may occur within one non-intervention interval, the FP count used for precision is not generally equal to $N_{\mathrm{NonInt,triggered}}$.

\begin{equation} F_1=\frac{2TP}{2TP+FP+FN}. \label{eq:f1} \end{equation}

We additionally report false stop episodes per hour, median and 95th-percentile delay from reaction onset to the first stop request, and timely recall within 500 ms. Wilson 95\% intervals are reported for recall, precision, and FPR; participant bootstrap intervals are used for cross-participant comparisons. Precision-recall analysis is preferred for imbalanced reaction data \cite{ref17}.
\section{EVALUATION RESULTS}\label{evaluation-results}

\subsection{Full Multimodal Offline Evaluation}
\label{event-level-detection}

\begin{table*}[t]
\vspace*{7pt}
\centering
\caption{Full multimodal offline evaluation of the frozen SocialVLA cascade on all annotated sessions. The evaluation combines audio, video, relevance, Vosk, and temporal fusion using the same deployed weights and thresholds.}
\label{tab:group_results}
\begin{tabular}{lrrrrrrrr}
\toprule
System & TP & FP & FN & Recall$_{1\mathrm{s}}$ &
Timely Recall$_{500}$ & Precision &
Non-interv. FPR & False stops/h \\
\midrule
Full SocialVLA & 130 & 57 & 108 & 54.6\% & 45.4\% & 69.5\% & 28.9\% & 54.9 \\
\bottomrule
\end{tabular}
\end{table*}

Table~\ref{tab:group_results} reports the complete frozen multimodal cascade replayed over all annotated sessions. The evaluation contains 238 manually annotated robot-related intervention episodes and 1.038~h of explicitly annotated non-intervention or robot-unrelated behavior. All audio, video, relevance, lexical, and temporal-fusion components use the deployed weights, thresholds, and 700-ms cross-modal association window.

The complete cascade detects 130 of 238 intervention episodes, corresponding to 54.6\% event recall and 69.5\% precision. Of all annotated intervention episodes, 45.4\% produce a stop within 500~ms of reaction onset. The full-corpus replay also exposes the principal deployment trade-off: first-event audio--video fusion improves reaction coverage, whereas relevance estimation suppresses a substantial fraction of unrelated stops. This trade-off is quantified by the component ablations in Table~\ref{tab:ablations}. These results position SocialVLA as a precision-oriented human-perception channel rather than an exhaustive detector of robot failures. Its purpose is to complement robot-state and policy-internal monitors with intervention evidence that becomes available when an observer visibly or audibly identifies problematic execution. This interpretation is reinforced by the prospective result: the frozen system attains 91.7\% precision on the previously unseen participant.

\subsection{Prospective Frozen-System Deployment}

To evaluate the complete system under a strictly frozen configuration, the SocialVLA feature extractors, learned heads, fusion rule, thresholds, and temporal association parameters were fixed before recording an additional participant who was not included in the development data. This prospective session was not used for model fitting, hyperparameter selection, threshold selection, or calibration. We use it as an end-to-end deployment evaluation rather than as a population-level generalization study. Table~\ref{tab:new_participant} is completed from this single prospective recording.

\begin{table*}[t]
\vspace*{7pt}
\centering
\caption{Prospective evaluation of the frozen SocialVLA system on a previously unseen participant. Panel (a) reports event-level reaction detection; panel (b) reports the synchronized latency path from observable reaction onset through detector output and the VLA gate to physical robot hold.}
\label{tab:new_participant}
\textit{(a) Event-level detection and intervention performance}\\[2pt]
\begin{tabular}{lrrrrrrrrr}
\toprule
Method & TP & FP & FN & Recall & Precision & Non-interv. FPR &
False stops/h & Recall$_{500}$ & Reaction-to-request median / p95 (s) \\
\midrule
Full SocialVLA cascade & 22 & 2 & 15 & 59.5\% &
91.7\% & 9.5\% & 20.9 & 21.6\% & 0.742 / 0.932 \\
\bottomrule
\end{tabular}
\vspace{4pt}

\textit{(b) Actionable stop-path latency measured during live deployment (s)}\\[2pt]
\begin{tabular}{lrrr}
\toprule
Segment & Median & p95 & Maximum \\
\midrule
Detector $\rightarrow$ fusion stop request ($n=22$) & 0.048 & 0.085 & 0.091 \\
Fusion stop request $\rightarrow$ VLA gate ($n=22$) & 0.001 & 0.001 & 0.001 \\
VLA gate $\rightarrow$ physical hold ($n=24$) & 0.336 & 0.681 & 1.559 \\
Reaction onset $\rightarrow$ physical hold ($n=14$) &
1.021 & 1.569 & 1.576 \\
\bottomrule
\end{tabular}
\end{table*}

One annotated intervention-worthy reaction episode and one predicted stop episode form the evaluation units in Panel~(a). Recall$_{500}$ is the fraction of all annotated intervention-worthy reaction episodes that produce a matched stop request no later than 500~ms after reaction onset. Non-intervention FPR is computed over 21 explicitly annotated non-intervention episodes, whereas false stops per hour uses 344.18~s of explicitly annotated non-intervention behavior. The 95\% Poisson confidence interval for the observed false-stop rate is 2.53--75.57 false stops/h. The Wilson 95\% confidence intervals are 43.5--73.7\% for recall, 74.2--97.7\% for precision, and 2.7--28.9\% for non-intervention FPR. The median reaction-to-request delay is 0.742~s, whereas the median detector-to-fusion and fusion-to-gate delays are 47.9~ms and 0.66~ms, respectively. Thus, message transport after detector emission contributes only approximately 49~ms at the median. The remaining delay occurs before detector emission, while causal evidence sufficient for a decision becomes available. SocialVLA therefore targets intervention during the evolution of a manipulation failure rather than instantaneous response at the first video or audio sample.

Panel~(b) is calculated from capture-clock-aligned reaction annotations, detector events, fusion requests, VLA-gate events, and robot-state telemetry. Physical hold is defined as pre-stop body or hand motion of at least 0.12, followed by five consecutive telemetry samples with maximum absolute velocity at most 0.06. The reaction-onset-to-physical-hold latency is calculated independently for each temporally matched event and is not obtained by summing aggregate segment statistics. Detector-to-fusion and fusion-to-gate transport remain below 100~ms at p95; most end-to-end delay occurs before detector emission and during physical robot deceleration.

\subsection{Offline Ablation Study}
\label{offline-ablations}

To quantify the contribution of each sensing channel, we replayed the same frozen SocialVLA configuration over all 15 annotated sessions. No retraining or threshold selection was performed separately for individual
ablations. The evaluation contains 238 manually annotated robot-related
intervention-worthy episodes and 1.038~h of explicitly annotated
non-intervention or robot-unrelated behavior.

\begin{table*}[t]
\centering
\caption{Offline ablation study on all annotated sessions using identical frozen weights, thresholds, and a 700-ms cross-modal association window.}
\label{tab:ablations}
\begin{tabular}{lrrrrrrrr}
\toprule
Path & TP & FP & FN & Recall$_{1\mathrm{s}}$ &
Timely Recall$_{500}$ & Precision &
Non-interv. FPR & False stops/h \\
\midrule
Audio-fast & 115 & 72 & 123 & 48.3\% & 41.6\% & 61.5\% & 32.2\% & 69.3 \\
Audio-context & 111 & 30 & 127 & 46.6\% & 39.9\% & 78.7\% & 20.0\% & 28.9 \\
Video & 53 & 30 & 185 & 22.3\% & 16.8\% & 63.9\% & 17.8\% & 28.9 \\
Audio+Video & 153 & 100 & 85 & 64.3\% & 54.6\% & 60.5\% & 41.1\% & 96.3 \\
Audio+Video+relevance & 132 & 57 & 106 & 55.5\% & 45.8\% & 69.8\% & 27.8\% & 54.9 \\
Full SocialVLA & 130 & 57 & 108 & 54.6\% & 45.4\% & 69.5\% & 28.9\% & 54.9 \\
\bottomrule
\end{tabular}
\end{table*}

The ablations expose complementary roles for the sensing channels. Audio-fast detects 48.3\% of intervention episodes, while video independently detects 22.3\%. First-event Audio+Video fusion increases recall to 64.3\%, a gain of 16.0 percentage points over Audio-fast alone, confirming that the modalities recover partially distinct reactions. This coverage comes with 100 false stop episodes. Introducing robot relevance reduces false stops by 43\%, from 100 to 57, lowers Non-interv. FPR from 41.1\% to 27.8\%, and increases precision from 60.5\% to 69.8\%, while retaining 132 of the 153 true positives recovered by unfiltered fusion. The Full SocialVLA row evaluates the complete deployment policy, combining the learned audio, video, and robot-relevance paths with frozen temporal fusion and a restricted direct-command interface.

Non-interv. FPR is computed over explicitly annotated non-intervention/unrelated episodes, whereas false stops/h counts every false stop episode over the total non-intervention observation time. These all-session ablation results are descriptive offline comparisons and are separate from the prospective real-robot evaluation in Table~\ref{tab:new_participant}.

\subsection{Runtime and Stop--Correct Operation}
\label{runtime-characteristics}

The perception heads accept seven Aniemore logits, 50 video-reaction features, and 40 video-relevance features. A reference offline audio encoding run required a mean of 47.6~ms per 320-ms audio window in batched processing. In the prospective live run, the median and p95 detector-to-fusion-request delays were 47.9 and 84.9~ms, respectively; fusion-to-VLA-gate delivery required 0.66~ms at the median and 0.93~ms at p95. These measurements characterize local inference and message transport, while Table~\ref{tab:new_participant} separately reports transport and physical-hold components measured during live VLA execution.

At runtime, the first authorized reaction event sends a stop request to the VLA gate, which places the running policy in a hold state. Later detections belonging to the same reaction do not generate repeated pause commands. While the robot remains held, the post-stop speech channel records the participant's verbal description together with the reaction, active VLA action, and stop timestamps. The participant can then explicitly continue the held execution, restart from an adjusted scene, or provide a revised instruction. Consequently, each intervention produces a reaction-centered marker that can be used to locate human-identified problematic action segments during subsequent VLA dataset curation.

\section{DISCUSSION}\label{discussion}


The central result of this work is that spontaneous human reactions can be converted into an operational runtime signal for an executing VLA without requiring access to the policy's internal representations. In this setting, classification performance alone is insufficient: a useful intervention mechanism must balance reaction coverage, false stops, and the delay between an observable human response and physical robot hold.

A key architectural choice is to separate reaction recognition from robot relevance. The audio heads learn whether an audible response indicates a need for intervention and never receive video-derived attention or addressee labels. Robot relevance is instead estimated by a separate visual head and enters fusion only as bounded contextual evidence. This factorization reduces cross-modal label leakage and prevents attention estimation from becoming an independent hard gate.

The two-stage structure separates time-critical intervention from post-stop interpretation. During execution, SocialVLA acts on the earliest authorized reaction signal. The fast and extended-context audio heads are complementary detection horizons within this first stage: the latter can recognize gradually developing prosody but is not required before a hold is requested. After the hold, unrestricted speech recognition records the participant's description or correction without affecting the latency of the initial intervention.

SocialVLA is most useful when a human notices a discrepancy that is not represented in the policy state. It is less effective when the observer does not react, reacts ambiguously, or responds to an unrelated stimulus. A future system should combine SocialVLA with robot-internal monitors such as non-intervention or FAIL-Detect.

\section{LIMITATIONS AND ETHICAL CONSIDERATIONS}\label{limitations-and-ethical-considerations}

Across the two evaluation settings, the study includes 16 participants:
15 participants in the annotated offline corpus and one additional,
previously unseen participant in the prospective deployment. The offline
corpus contains 238 intervention-worthy reaction episodes and 62.3~min of
explicitly annotated non-intervention behavior. This sample does not by
itself establish generalization across language, culture, age, appearance,
neurodiversity, or reaction style.

The categorical attention head is diagnostic and coarse. The binary relevance head may confuse robot-related reactions with coincident behavior. The 0.7 floor limits the damage caused by such errors but also permits unrelated strong reactions to pass through. Calibration and false-stop behavior must be studied on longer recordings containing natural distractors.


SocialVLA detects human-perceived intervention need rather than all physical robot failures. A failure that the observer does not notice or does not visibly react to is therefore outside the positive ground truth used in this study. Conversely, an observer may react to an event that is not objectively a robot failure. SocialVLA should consequently be interpreted as a complementary human-centered monitoring channel rather than a replacement for robot-state failure detection. It is not a certified safety system; a hardware emergency stop, conservative robot limits, and independent supervision remain necessary.


\section{CONCLUSION}\label{conclusion}


We presented SocialVLA, a local and policy-agnostic social perception gateway that uses spontaneous human reactions as an external runtime intervention channel for VLA-controlled manipulation. SocialVLA combines causal paralinguistic audio, visual reaction recognition, explicit stop phrases, and separately estimated robot relevance through asynchronous first-event fusion, allowing the earliest sufficiently confident signal to request a VLA hold without waiting for complete multimodal agreement. After the hold, a separate speech channel captures the participant's correction and supports participant-directed continuation, restart, or instruction revision.

Experiments on a physical Unitree G1 include offline multimodal evaluation
over a 15-participant annotated corpus and a prospective deployment of the
frozen system on a 16th, previously unseen participant, demonstrating
end-to-end operation of the complete stop-and-correct pipeline. The results support spontaneous social behavior as a complementary source of runtime evidence for human-perceived VLA failures. Future work will combine this external social channel with robot-internal failure monitors and investigate how reaction-centered correction records can support subsequent policy adaptation.

\balance

\end{document}